\documentclass[runningheads]{llncs}
\usepackage[T1]{fontenc}
\usepackage{amssymb}
\usepackage{amsmath}
\usepackage{multirow}
\usepackage{booktabs}
\usepackage{float}
\usepackage{graphicx}
\begin{document}
\title{AA-ViT: Anatomically Aware Vision Transformer with Structural and Frequency Guidance for Contrast Enhanced Brain MRI Synthesis}
%
%

\author{
Talha Meraj\inst{1,2,3,8}\and
Tom Flannery\inst{4}\and
Charlie Cummins\inst{5}\and
Matt Townend\inst{6}\and
Thomas C Booth\inst{5,6}\and
Peter Crossley\inst{4}\and
Michael McCann\inst{3,7}\and
Ian Overton\inst{8}\and
Saritha Unnikrishnan\inst{1,2,3,*}
}

\authorrunning{T. Meraj et al.}

\institute{
Department of Computing and Electronic Engineering, Atlantic Technological University, Sligo, Ireland \and
Mathematical Modelling and Intelligent Systems for Health and Environment (MISHE), Faculty of Engineering and Design, Atlantic Technological University, Sligo, Ireland \and
JANUS Research Centre, Atlantic Technological University, Letterkenny, Ireland \and
Department of Neurosurgery, Belfast Health and Social Care Trust, Belfast, BT12 6BA, UK \and
Department of Neuroradiology, Ruskin Wing, King's College Hospital NHS Foundation Trust, London, SE5 9RS, UK \and
School of Biomedical Engineering \& Imaging Sciences, King's College London, UK \and
Department of Computing, Atlantic Technological University, Letterkenny, Ireland \and
Johnston Cancer Research Centre, Queen’s University Belfast, Belfast, BT9 7AE, UK
}

\maketitle              
\begin{abstract}
Accurate tumour localization and diagnosis is a critical component of clinical care for brain cancers. Magnetic Resonance Imaging (MRI) is the most commonly used imaging modality due to its superior soft-tissue contrast. However, standard MRI often exhibits limited contrast and imaging artifacts, which necessitates the use of contrast agents to enhance lesion visibility. The administration of chemical contrast agents is not always feasible and may be contraindicated in patients with renal impairment or other health conditions. As a result, developing accurate and non-invasive contrast enhanced MRI (CEMRI) synthesis methods has clinical importance. In recent years, numerous approaches for CEMRI synthesis have been proposed, predominantly relying on generative artificial intelligence models. While these methods demonstrate promising performance, their dependence on implicit feature learning often limits their ability to preserve anatomical boundaries and tumour-specific fine structures. To address these challenges, we propose an anatomically aware frequency-and-structure-guided vision transformer (AA-ViT), for CEMRI synthesis using pre-contrast MRI modalities (T1, T2, and FLAIR). Experiments on the BraTS 2021 dataset demonstrate that the proposed method preserves anatomical and lesion boundaries, achieving higher PSNR and SSIM than state-of-the-art approaches. Clinical evaluation by three neuroradiologists and a neurosurgeon on 19 randomly selected cases across diverse gliomas yielded a mean score of 3.94/5, providing preliminary clinical validation rarely seen in prior studies. Synthetic post-contrast scans from our model could lower scanning costs, shorten imaging time, and avoid the potential risks of using gadolinium-based contrast agents.

\end{abstract}
\section{Introduction}
Contrast enhanced magnetic resonance images (CEMRI) are indispensable for brain tumour evaluation and treatment planning \cite{ghaffari2019automated,villanueva2017current}. It provides high-resolution information on tumour boundaries and differentiation between active and necrotic regions \cite{chang2025controllable,yang2024segmentation}. Gadolinium-based contrast agents (GBCAs) are used in CEMRI which poses safety concerns such as potential deposition in the brain \cite{gulani2017gadolinium}, allergic reactions and nephrogenic system fibrosis \cite{zhang2023synthesis}, in addition to increasing scan time and cost \cite{pang2025d}. The European Medicines Agency restricts the use of GBCAs while permitting continued use of safer macrocyclic agents \cite{Kleesiek2019VirtualContrast}. Given this, eliminating contrast agents while preserving diagnostically relevant image contrast is highly desirable \cite{pang2025d}. To reduce the usage of GBCAs in brain cancer treatment, recent research has focused on synthesizing CEMRIs from non-contrast MR images (T1, T2-weighted, and FLAIR scans) using advanced generative models such as generative adversarial methods (GANs) \cite{li2019diamondgan,yurt2022progressively}, diffusion models \cite{ozbey2023unsupervised,pang2025d}, and flow-matching \cite{chang2025controllable} approaches. MRI synthesis can also play a pivotal role in the case of a missing MRI modality. However, generating single modality from one available modality loses the essential tissue related information, present in other non-contrast MR images causing difficulty in the diagnosis process. Therefore, multimodal MRI synthesis has received attention in recent research \cite{dayarathna2025mu}.

For multi-modal MRI synthesis, prior studies have improved contextual guidance and fusion strategies using transformers, state-space models (SSMs), and diffusion models. ResVit \cite{dalmaz2022resvit} integrates convolutional local features with transformer-based global context, achieving strong performance, but its reliance on implicit features limits precise preservation of anatomical and tumor boundaries. Likewise, \cite{han2023explainable} focused on adaptive weighting and interpretability across multi-input sequences but did not explicitly constrain anatomical consistency in the generated images. More recent architectures such as I2I-Mamba \cite{atli2024i2i}, capture long-range dependencies through state-space modeling, improving global coherence, yet still lack explicit spatial constraints for anatomical edges and lesion boundaries. Diffusion-based frameworks such as MU-Diff \cite{dayarathna2025mu} enhance lesion realism via mutual learning and uncertainty modeling, generating high-resolution images but with increased computational cost and prolonged inference time, limiting clinical applicability.
\par To provide a solution on aforementioned problems, we enhance a ResViT-based image synthesis framework by introducing an anatomically aware vision transformer (AA-ViT) which incorporates (1) improved encoder blocks as residual dense edge block (RDEB), an edge-aware residual dense encoding. (2) Additionally, we introduce self-error focusing, anatomically guided, and frequency-domain loss components in image generator to enhance structural fidelity and lesion preservation. These complementary objectives improve boundary precision and high-frequency detail recovery, enabling more accurate and efficient brain MRI synthesis that outperforms SOTA transformer, SSM, and diffusion-based models.

\section{Methods}
\begin{figure}
\includegraphics[width=\textwidth]{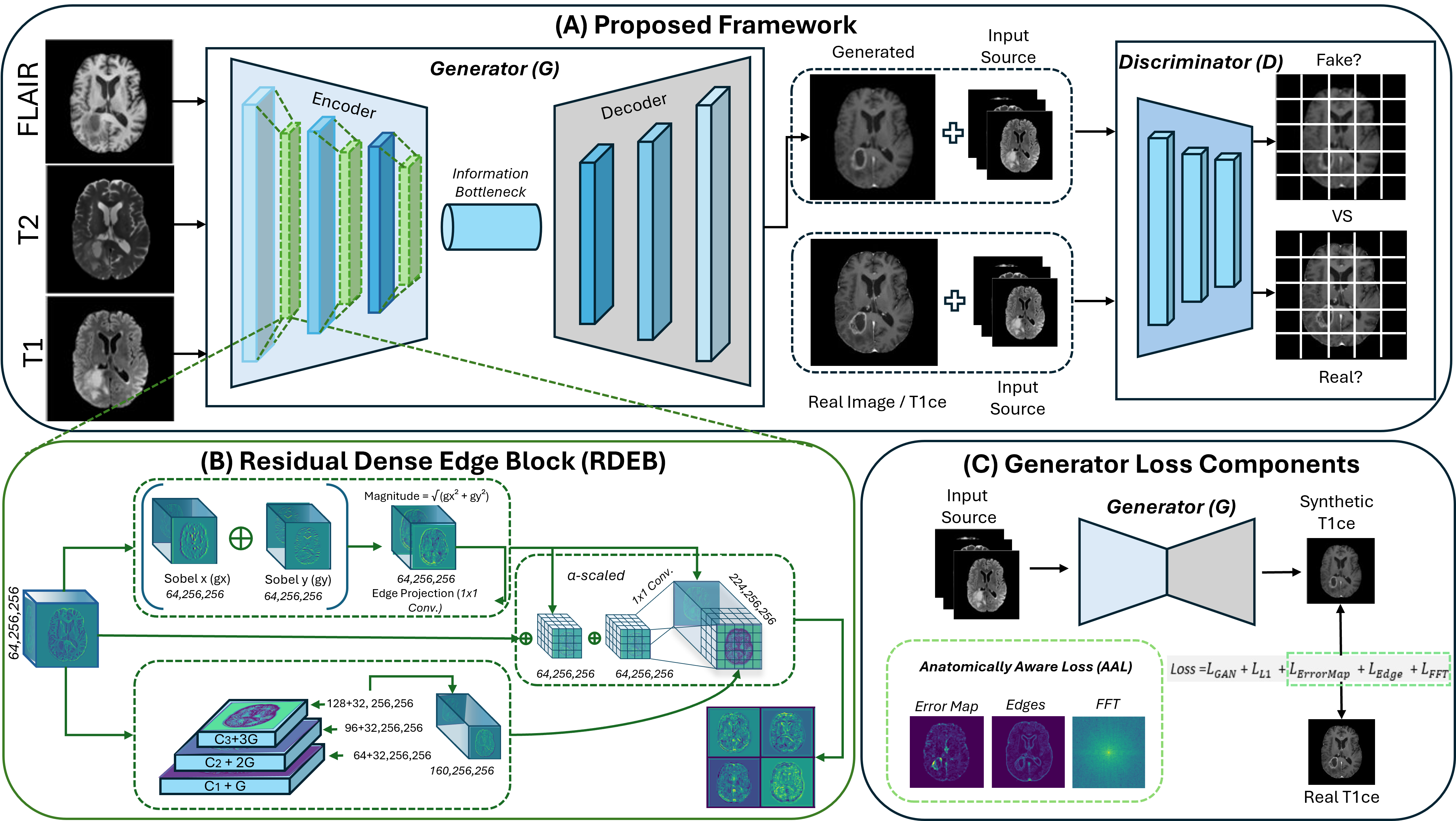}
\caption{Overview of the proposed AA-ViT framework. 
(A) Overall architecture for synthesizing contrast-enhanced MRI from pre-contrast inputs using a generator–discriminator setup. 
(B) Residual Dense Edge Block (RDEB) integrated within the encoder to enhance structural feature representation. 
(C) Multi-component anatomically aware training objective used to guide image synthesis.} \label{fig1}
\end{figure}

Fig. \ref{fig1} presents the overall architecture of the proposed AA-ViT framework. It follows a PatchGAN-based image-to-image translation paradigm, where the generator \textit{(G)} learns to synthesize contrast-enhanced MRI (CEMRI) from pre-contrast MRI inputs. The generator incorporates an encoder enhanced with a Residual Dense Edge Block (RDEB) to improve feature representation using gradient-based edge information. A patch-based discriminator \textit{(D)} evaluates local image realism to guide the generator toward producing structurally consistent outputs. Additionally, anatomically aware supervision is enforced through multiple loss components that promote reconstruction fidelity, edge preservation, and frequency consistency.

\subsection{Anatomically Aware Encoder with Residual Dense Edge Block (RDEB)}
The proposed AA-ViT enhances the encoder with anatomical edge priors while retaining the ResViT \cite{dalmaz2022resvit} transformer for global context. It consists of a generator $G$ that converts pre-contrast MRI into contrast-enhanced images, and a PatchGAN discriminator $D$ that ensures patch-level realism. Multi-modal MRI scans (T1, T2, FLAIR) are treated as RGB channels and fed into the encoder $E$, producing feature maps $\mathbf{F}_0 \in \mathbb{R}^{C \times H \times W}$, where $\mathbf{F}_0$ is the encoded feature tensor,  $C$ is the number of channels, and $H,W$ are spatial dimensions. These features form the basis for dense feature extraction, with each layer building new patterns while reusing all previous outputs (Eq. \ref{eq1}):

\begin{equation}
\mathbf{F}_l = \sigma\!\left(\mathcal{C}_l\!\left([\mathbf{F}_0, \mathbf{F}_1, \dots, \mathbf{F}_{l-1}]\right)\right), 
\quad l = 1, \dots, L,
\label{eq1}
\end{equation}

with $\mathcal{C}_l(\cdot)$ as a $3\times3$ convolution, $[\cdot]$ denoting channel-wise concatenation, and $\sigma(\cdot)$ as the ReLU activation. This dense connectivity not only stabilizes learning but also ensures no feature is wasted. To make the network aware of anatomical boundaries, edge information is extracted from the initial features using sobel kernels (Eq. \ref{eq2}):

\begin{equation}
\mathbf{E} = \sqrt{(\mathbf{F}_0 * \mathbf{S}_x)^2 + (\mathbf{F}_0 * \mathbf{S}_y)^2 + \epsilon},
\label{eq2}
\end{equation}

where $\mathbf{S}_x$ and $\mathbf{S}_y$ capture horizontal and vertical edges, and $*$ represents convolution. These edge maps highlight structures that are important for medical interpretation. Finally, the dense features and projected edge information are fused and combined with the residual connection to form the output of the block (Eq. \ref{eq3}):

\begin{equation}
\mathbf{F}_{\text{out}} = \mathbf{F}_0 
+ \mathcal{C}_{1\times1}\!\left([\mathbf{F}_0, \mathbf{F}_1, \dots, \mathbf{F}_L, \mathcal{P}(\mathbf{E})]\right)
+ \alpha\,\mathcal{P}(\mathbf{E}),
\label{eq3}
\end{equation}

where $\mathcal{P}(\mathbf{E})$ is a learnable projection of the edge features, and $\alpha$ adjusts their influence. In this way, the network naturally integrates anatomical priors with multi-layer feature learning into Aggregated Residual Transformer (ART) blocks \cite{dalmaz2022resvit}, producing rich, edge-aware representations that are ready for downstream contrast enhancement tasks. However, to justify the contribution of the proposed RDE blocks, we conducted a layer-wise representation similarity analysis using centered kernel alignment (CKA). We compare features extracted from synthesized and real CEMRI images for both the baseline ResViT and the proposed AA-ViT framework. 

\begin{figure}
\includegraphics[width=\textwidth]{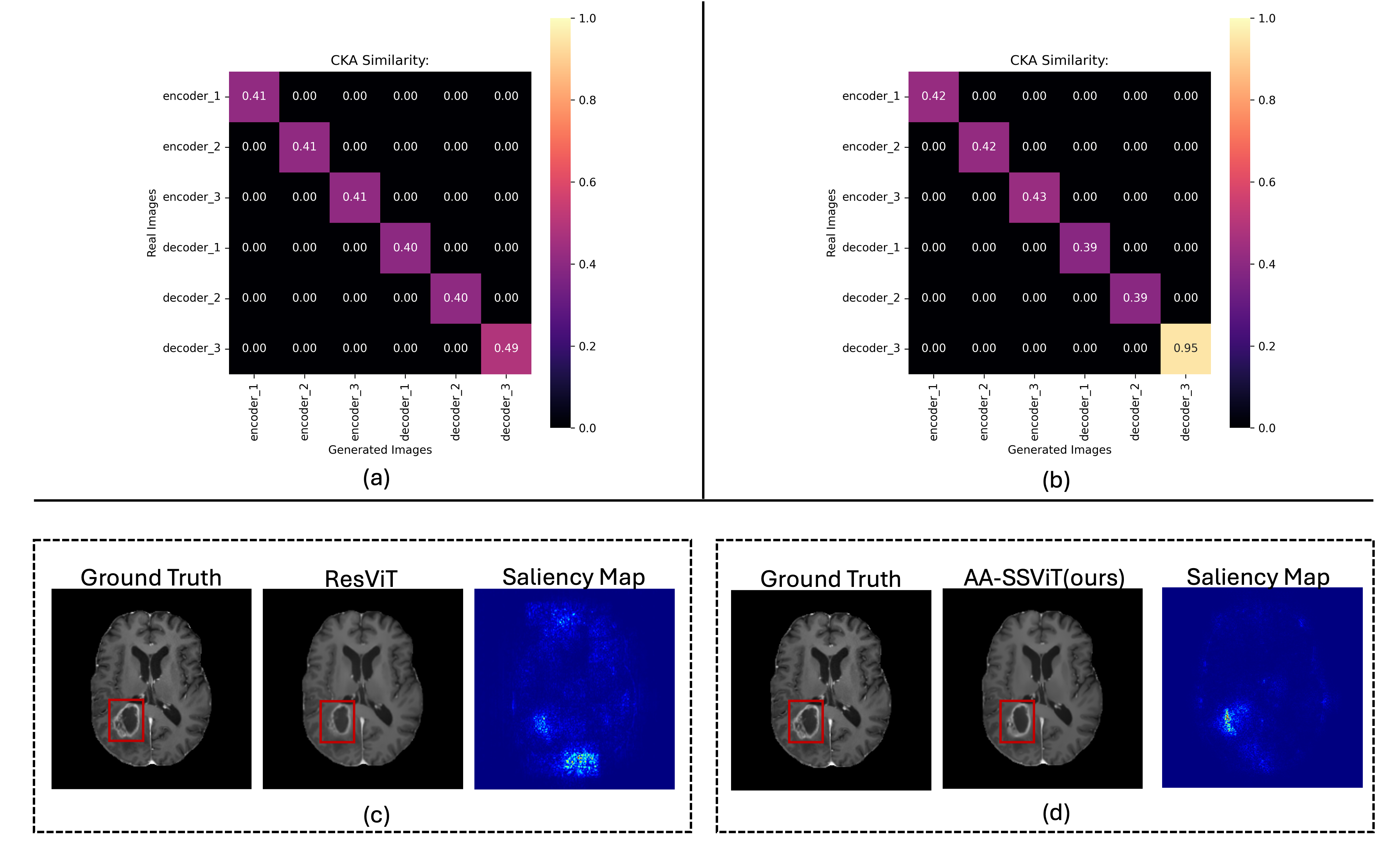}
\caption{Layer-wise CKA comparison between features extracted from synthesized and real CEMRI images for (a) ResViT and (b) AA-ViT (ours). Saliency comparison: (c) ResViT attends to irrelevant regions, whereas (d) AA-ViT focuses on anatomically important areas needing improvement.} \label{fig2}
\end{figure}

As shown in Fig. \ref{fig2} section (a) and (b), AA-ViT consistently achieves higher CKA \cite{kornblith2019similarity} scores across all encoder layers, indicating improved anatomical feature alignment enabled by explicit edge-aware dense encoding. Notably, the final decoder layer of AA-ViT exhibits a substantial increase in CKA (0.948 ± 0.102) compared to ResViT (0.486 ± 0.311), demonstrating that the proposed architectural and loss modifications effectively align synthesized representations with real CEMRI anatomy. These results provide strong representation-level evidence that RDEB enhances anatomical fidelity beyond visual inspection and pixel-wise metrics.

\subsection{Anatomically Aware Loss with Structural and Frequency Guidance (AAL)}
After improving the generator $(G)$, we introduce an anatomically-aware loss to enforce intensity fidelity, anatomical boundaries, and high-frequency detail reconstruction, defined in Eq.~\ref{eq4}:

\begin{equation}
\mathcal{L}_{\text{AAL}} =
\lambda_1 \mathcal{L}_{\text{L1}} +
\lambda_2 \mathcal{L}_{\text{adv}} +
\lambda_3 \mathcal{L}_{\text{err}} +
\lambda_4 \mathcal{L}_{\text{edge}} +
\lambda_5 \mathcal{L}_{\text{FFT}}.
\label{eq4}
\end{equation}

where $\lambda_i$ are weighting coefficients. Here, $\mathcal{L}*L1$ ensures pixel-wise fidelity and $\mathcal{L}$*adv promotes realism. The error focused term $\mathcal{L}_{\text{err}}$ targets regions with high reconstruction error, calculated as in (Eq. \ref{eq5}) :

\begin{equation}
\mathcal{L}_{\text{err}} = \frac{1}{HW} \sum_{i,j} \big| \mathbf{Y}_{i,j} - \hat{\mathbf{Y}}_{i,j} \big|
\label{eq5}
\end{equation}

where $\mathbf{Y}$ and $\hat{\mathbf{Y}}$ are the ground-truth and generated images, and $H$, $W$ are the image dimensions. To preserve anatomical boundaries, the edge loss $\mathcal{L}_{\text{edge}}$ is defined as (Eq. \ref{eq6}):

\begin{equation}
\mathcal{L}_{\text{edge}} = 
\Big\| \text{$C_l$}(\hat{\mathbf{Y}}, S_x) - \text{$C_l$}(\mathbf{Y}, S_x) \Big\|_1 \;+\; 
\Big\| \text{$C_l$}(\hat{\mathbf{Y}}, S_y) - \text{$C_l$}(\mathbf{Y}, S_y) \Big\|_1
\label{eq6}
\end{equation}

where $S_x,S_y$ are Sobel kernels and $C(\cdot, S)$ denotes convolution. High-frequency details are enforced using the FFT loss $\mathcal{L}_{\text{FFT}}$, as in Eq.~\ref{eq7}:

\begin{equation}
\mathcal{L}_{\text{FFT}} = 
\frac{1}{HW} \sum_{i,j} 
\big| \mathcal{F}(\hat{\mathbf{Y}})_{i,j} - \mathcal{F}(\mathbf{Y})_{i,j} \big| \cdot M_{i,j}
\label{eq7}
\end{equation}

with the radial high-pass mask $M_{i,j}$ defined in Eq.~\ref{eq8}:

\begin{equation}
M_{i,j} = 
\begin{cases} 
1, & r(i,j) > r_{\text{frac}} \\
0, & \text{otherwise}
\end{cases}, 
\quad
r(i,j) = \frac{\sqrt{(i-H/2)^2 + (j-W/2)^2}}{\max(r)}
\label{eq8}
\end{equation}

Here, $\mathcal{F}(\cdot)$ is the 2D Fourier transform and $r_{\text{frac}}$ defines the high-frequency threshold. Together, $\mathcal{L}_{\text{AAL}}$ enforces intensity fidelity, perceptual realism, anatomical consistency, and fine-detail reconstruction, reducing hallucinations and emphasizing relevant errors over baseline losses (Fig.~\ref{fig2}c–d).

\textbf{Clinical Validation}
Evaluation by clinicians was performed by three neuroradiologists and one neurosurgeon on 19 randomly selected cases, comprising 1,316 images representing a diverse set of gliomas (normal, solitary, multi-focal, and multicentric). Clinicians independently scored the generated post-contrast images on a 5-point Likert scale (1 = lowest, 5 = highest) across five questions: Q1) image quality, Q2) diagnostic utility, Q3) lesion visibility and clarity, Q4) confidence in clinical use, and Q5) whether the images provided additional information beyond standard T1, T2, and FLAIR sequences. 

\section{Experiments and Results}
\subsection{Dataset and Baselines}
We used the BraTS2021 dataset \cite{Baid2021BraTS} with 1,251 subjects and four aligned MRI sequences (T1, T1ce, T2, FLAIR), splitting 830/93/328 for training, validation, and testing following to TSF-Seq2Seq \cite{han2023explainable}; axial slices were scored by intensity across non-zero voxels, retaining the top half ($\le 100$ per subject), resized to $256\times256$, normalized to [0,1], and stacked into a four-channel input for AA-ViT (first three channels: pre-contrast, fourth: T1ce ground truth).

\textbf{Implementation Details}
Training used the Adam optimizer ($\beta_1=0.5$) with a batch size of 24 and learning rate $2\times10^{-4}$. The total loss combined adversarial, reconstruction, perceptual, edge, multi-scale, difference-map, and frequency-domain components with empirically chosen weights. Experiments were run on an NVIDIA A100 GPU (80 GB VRAM) with CUDA 12.1.

\begin{table}[t]
\centering
\caption{Performance comparison on the BraTS2021 dataset. PSNR (dB) and SSIM are reported as mean $\pm$ standard deviation. Best results are highlighted in bold.}
\label{tab:1}
\begin{tabular}{lcc}
\hline
\textbf{Model} & \textbf{PSNR (dB)} $\uparrow$ & \textbf{SSIM} $\uparrow$ \\
\hline
ResViT\cite{dalmaz2022resvit}
& $26.919 \pm 4.657$ 
& $0.916 \pm 0.040$ \\

I2I Mamba \cite{atli2024i2i}
& $26.615 \pm 4.195$ 
& $0.903 \pm 0.038$ \\

TSF-Seq2Seq \cite{han2023explainable}
& $26.681 \pm 1.908$ 
& $0.847 \pm 0.041$ \\

MU-Diff \cite{dayarathna2025mu}
& $24.218 \pm 3.576$ 
& $0.895 \pm 0.040$ \\

\textbf{AA-ViT (Ours)} 
& $\mathbf{27.790 \pm 4.927}$ 
& $\mathbf{0.930 \pm 0.039}$ \\
\hline
\end{tabular}
\end{table}

\textbf{Quantitative Results }
In Table \ref{tab:1}, we compared AA-ViT with four SOTA multi-modal MRI synthesis methods from different families: ResViT \cite{dalmaz2022resvit}, TSF-seq2seq \cite{han2023explainable}, I2I-Mamba \cite{atli2024i2i}, and MU-Diff \cite{dayarathna2025mu}. Table \ref{tab:1} shows performance on BraTS2021 using PSNR and SSIM. AA-ViT achieves the highest scores (PSNR $27.71 \pm 4.91$ dB, SSIM $0.929 \pm 0.039$), demonstrating improved reconstruction fidelity and structural preservation. While I2I-Mamba and TSF-seq2seq have comparable PSNR, their lower SSIM ($0.903$, $0.847$) indicates less accurate anatomical detail. MU-Diff shows moderate performance (PSNR $24.22$ dB, SSIM $0.895$), reflecting limited cross-modal capture. These results confirm that AA-ViT effectively leverages anatomical priors for robust, structurally consistent CEMRI synthesis despite inter-subject variability and lesion heterogeneity.

\begin{figure}
\includegraphics[width=\textwidth]{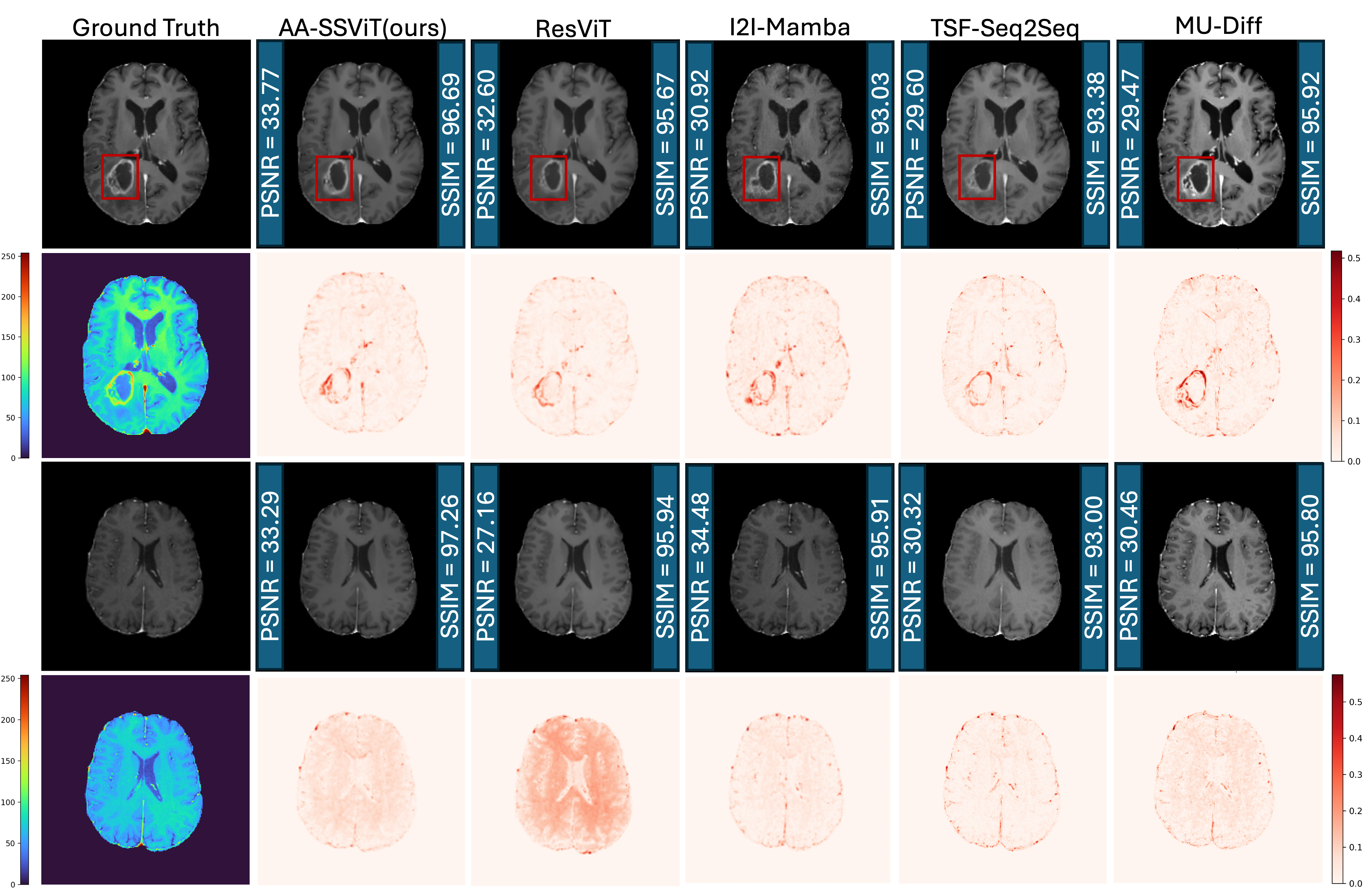}
\caption{Visualization of cancerous and normal synthesized images with PSNR and SSIM of AA-ViT and baselines (row 1 and 3) and corresponding error maps (row 2 and 4).} \label{fig3}
\end{figure}

\textbf{Qualitative Results }
Qualitative results of two samples including cancerous and normal slices shown in the first and third rows in Fig. \ref{fig3} showcase the T1ce synthesis results on the BRATS 2021 with improved PSNR and SSIM compared to other baselines. The second and fourth rows show the error maps in the form of heat maps, showing that AA-ViT can generate more anatomically and texturally fine images nearer to the original image compared to other baselines.

\subsection{Ablation Study and Preliminary Clinical Evaluation}
\textbf{Ablation Study} 
Baseline ResViT hyperparameters were first optimized. Components were then added sequentially to improve anatomical fidelity: error-map loss for weak-gradient regions, RDEB block with edge loss for boundaries, and FFT loss for fine details (Table~\ref{tab:2}). Error-map loss improves PSNR~27.519 and SSIM~0.926 by guiding reconstruction via residual errors, while RDEB with edge loss enhances structural consistency. The full model achieves the best performance: PSNR~27.790 and SSIM~0.930. Each component contributes positively, and all improvements are statistically significant.

\begin{table}[h!]
\centering
\caption{Ablation Study: PSNR and SSIM (mean ± SD) for each model component. Asterisks indicate significant improvement over baseline via Wilcoxon signed-rank test (*** $p<0.001$); best results in bold.}
\label{tab:2}
\scriptsize
\setlength{\tabcolsep}{4pt}
\renewcommand{\arraystretch}{1.2}

\begin{tabular}{lcc}
\hline
\textbf{Method} & \textbf{PSNR} $\uparrow$ & \textbf{SSIM} $\uparrow$ \\
\hline

Baseline 
& 26.919 ± 4.657 
& 0.916 ± 0.040 \\

+ $\mathcal{L}_{\text{err}}$
& 27.519 ± 4.655*** 
& 0.926 ± 0.037*** \\

+ $\mathcal{L}_{\text{err}}$ + RDEB + $\mathcal{L}_{\text{edge}}$ 
& 27.780 ± 4.866*** 
& 0.929 ± 0.039*** \\

+ $\mathcal{L}_{\text{err}}$ + RDEB + $\mathcal{L}_{\text{edge}}$ + $\mathcal{L}_{\text{FFT}}$
& \textbf{27.790 ± 4.927***} 
& \textbf{0.930 ± 0.039***} \\
\hline
\end{tabular}
\end{table}

\begin{figure}[H]
\includegraphics[width=\textwidth]{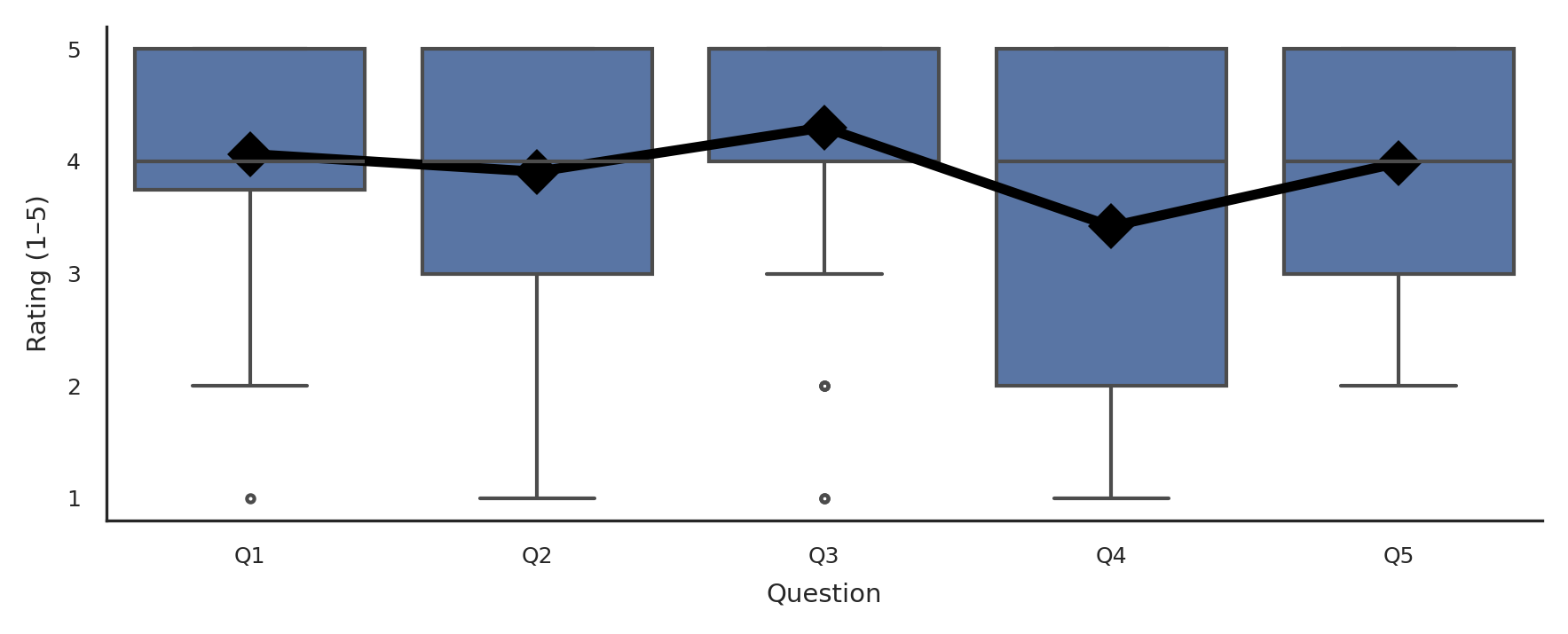}
\caption{Clinician ratings (Q1–Q5) of generated images: realism, diagnostic utility, lesion clarity, confidence, and extra information beyond T1/T2/FLAIR. Boxes = IQR \& median; whiskers = full range; points = outliers; black diamonds = mean.}
\label{fig4}
\end{figure}

Aggregated ratings (Fig. \ref{fig4}) showed high scores including for image quality, diagnostic utility, and lesion visibility. The mean overall score was 3.94/5, suggesting potential clinical value.

\section{Conclusion}
In this work, we proposed AA-ViT, an anatomically aware self-supervised vision transformer for synthesizing T1ce images from pre-contrast T1, T2, and FLAIR MRI. By integrating multi-modal information, using an edge-focused encoder, and leveraging error-map and edge-based supervision, AA-ViT achieves high-fidelity reconstruction. Experiments on BraTS2021 show it outperforms SOTA methods in PSNR and SSIM, and ablation studies highlight the contribution of each component. Evaluation of predictions by clinicians across 19 randomly sampled cases provides preliminary evidence supporting the value of AA-ViT across solitary, multi-focal, and multi-centric tumors. Limitations include empirically tuned hyperparameters and slice-wise processing; future work will explore context-aware strategies and clinician-guided optimization to enhance anatomical consistency.

\begin{credits}
\subsubsection{\ackname} This research is supported by TU RISE, which is co-funded by the Government of Ireland and the European Union through the European Regional Development Fund (ERDF) Southern, Eastern \& Midland Regional Programme 2021–2027 and the Northern \& Western Regional Programme 2021–2027. The authors also acknowledge the use of the JANUS High-Performance Computing (HPC) cluster at Atlantic Technological University (ATU) for providing computational resources for this work.

\subsubsection{\discintname}
The authors have no competing interests to declare that are
relevant to the content of this article.
\end{credits}

%
%
%
\bibliographystyle{splncs04}
\bibliography{mybib}

@article{yang2024segmentation,
  title={Segmentation method of magnetic resonance imaging brain tumor images based on improved UNet network},
  author={Yang, Yang and Wang, Peng and Yang, Zhenyu and Zeng, Yuecheng and Chen, Feng and Wang, Zhiyong and Rizzo, Stefania},
  journal={Translational Cancer Research},
  volume={13},
  number={3},
  pages={1567},
  year={2024}
}

@article{ghaffari2019automated,
  title={Automated brain tumor segmentation using multimodal brain scans: a survey based on models submitted to the BraTS 2012--2018 challenges},
  author={Ghaffari, Mina and Sowmya, Arcot and Oliver, Ruth},
  journal={IEEE reviews in biomedical engineering},
  volume={13},
  pages={156--168},
  year={2019},
  publisher={IEEE}
}

@article{villanueva2017current,
  title={Current clinical brain tumor imaging},
  author={Villanueva-Meyer, Javier E and Mabray, Marc C and Cha, Soonmee},
  journal={Neurosurgery},
  volume={81},
  number={3},
  pages={397--415},
  year={2017},
  publisher={LWW}
}

@article{gulani2017gadolinium,
  title={Gadolinium deposition in the brain: summary of evidence and recommendations},
  author={Gulani, Vikas and Calamante, Fernando and Shellock, Frank G and Kanal, Emanuel and Reeder, Scott B},
  journal={The Lancet Neurology},
  volume={16},
  number={7},
  pages={564--570},
  year={2017},
  publisher={Elsevier}
}

@article{zhang2023synthesis,
  title={Synthesis of contrast-enhanced breast mri using multi-b-value dwi-based hierarchical fusion network with attention mechanism},
  author={Zhang, Tianyu and Han, Luyi and D'Angelo, Anna and Wang, Xin and Gao, Yuan and Lu, Chunyao and Teuwen, Jonas and Beets-Tan, Regina and Tan, Tao and Mann, Ritse},
  journal={arXiv preprint arXiv:2307.00895},
  year={2023}
}

@inproceedings{pang2025d,
  title={D 3 M: Deformation-Driven Diffusion Model for Synthesis of Contrast-Enhanced MRI with Brain Tumors},
  author={Pang, Haowen and Zhang, Peng and Hong, Xiaoming and Chen, Shannan and Ye, Chuyang},
  booktitle={International Conference on Medical Image Computing and Computer-Assisted Intervention},
  pages={151--160},
  year={2025},
  organization={Springer}
}

@article{Kleesiek2019VirtualContrast,
  author  = {Kleesiek, Jens and Morshuis, Jan Nikolas and Isensee, Fabian and Deike-Hofmann, Katerina and Paech, Daniel and Kickingereder, Philipp and K{\"o}the, Ullrich and Rother, Carsten and Forsting, Michael and Wick, Wolfgang and Bendszus, Martin and Schlemmer, Heinz-Peter and Radbruch, Alexander},
  title   = {Can Virtual Contrast Enhancement in Brain MRI Replace Gadolinium?: A Feasibility Study},
  journal = {Investigative Radiology},
  volume  = {54},
  number  = {10},
  pages   = {653--660},
  year    = {2019},
  month   = {October},
  doi     = {10.1097/RLI.0000000000000583}
}

@inproceedings{chang2025controllable,
  title={Controllable Flow Matching for 3D Contrast-Enhanced Brain MRI Synthesis from Non-contrast Scans},
  author={Chang, Heng and Shang, Yu and Wang, Haifeng and Liang, Yuxia and Wang, Haoyu and Wang, Fan and Niu, Chen and Lian, Chunfeng},
  booktitle={International Conference on Medical Image Computing and Computer-Assisted Intervention},
  pages={119--128},
  year={2025},
  organization={Springer}
}

@inproceedings{li2019diamondgan,
  title={DiamondGAN: unified multi-modal generative adversarial networks for MRI sequences synthesis},
  author={Li, Hongwei and Paetzold, Johannes C and Sekuboyina, Anjany and Kofler, Florian and Zhang, Jianguo and Kirschke, Jan S and Wiestler, Benedikt and Menze, Bjoern},
  booktitle={International Conference on Medical Image Computing and Computer-Assisted Intervention},
  pages={795--803},
  year={2019},
  organization={Springer}
}

@article{yurt2022progressively,
  title={Progressively volumetrized deep generative models for data-efficient contextual learning of MR image recovery},
  author={Yurt, Mahmut and {\"O}zbey, Muzaffer and Dar, Salman UH and Tinaz, Berk and Oguz, Kader K and {\c{C}}ukur, Tolga},
  journal={Medical image analysis},
  volume={78},
  pages={102429},
  year={2022},
  publisher={Elsevier}
}

@article{ozbey2023unsupervised,
  title={Unsupervised medical image translation with adversarial diffusion models},
  author={{\"O}zbey, Muzaffer and Dalmaz, Onat and Dar, Salman UH and Bedel, Hasan A and {\"O}zturk, {\c{S}}aban and G{\"u}ng{\"o}r, Alper and Cukur, Tolga},
  journal={IEEE Transactions on Medical Imaging},
  volume={42},
  number={12},
  pages={3524--3539},
  year={2023},
  publisher={IEEE}
}

@article{dayarathna2025mu,
  title={MU-Diff: a mutual learning diffusion model for synthetic MRI with Application for brain lesions},
  author={Dayarathna, Sanuwani and Wu, Yicheng and Cai, Jianfei and Wong, Tien-Tsin and Law, Meng and Islam, Kh Tohidul and Peiris, Himashi and Chen, Zhaolin},
  journal={npj Artificial Intelligence},
  volume={1},
  number={1},
  pages={11},
  year={2025},
  publisher={Nature Publishing Group UK London}
}

@article{dalmaz2022resvit,
  title={ResViT: Residual vision transformers for multimodal medical image synthesis},
  author={Dalmaz, Onat and Yurt, Mahmut and {\c{C}}ukur, Tolga},
  journal={IEEE Transactions on Medical Imaging},
  volume={41},
  number={10},
  pages={2598--2614},
  year={2022},
  publisher={IEEE}
}

@inproceedings{han2023explainable,
  title={An explainable deep framework: towards task-specific fusion for multi-to-one MRI synthesis},
  author={Han, Luyi and Zhang, Tianyu and Huang, Yunzhi and Dou, Haoran and Wang, Xin and Gao, Yuan and Lu, Chunyao and Tan, Tao and Mann, Ritse},
  booktitle={International Conference on Medical Image Computing and Computer-Assisted Intervention},
  pages={45--55},
  year={2023},
  organization={Springer}
}

@article{atli2024i2i,
  title={I2I-Mamba: Multi-modal medical image synthesis via selective state space modeling},
  author={Atli, Omer F and Kabas, Bilal and Arslan, Fuat and Demirtas, Arda C and Yurt, Mahmut and Dalmaz, Onat and Cukur, Tolga},
  journal={arXiv preprint arXiv:2405.14022},
  year={2024}
}

@inproceedings{kornblith2019similarity,
  title={Similarity of neural network representations revisited},
  author={Kornblith, Simon and Norouzi, Mohammad and Lee, Honglak and Hinton, Geoffrey},
  booktitle={International conference on machine learning},
  pages={3519--3529},
  year={2019},
  organization={PMlR}
}

@article{Baid2021BraTS,
  title   = {The {RSNA-ASNR-MICCAI} {BraTS} 2021 Benchmark on Brain Tumor Segmentation and Radiogenomic Classification},
  author  = {Baid, U. and Ghodasara, S. and Mohan, S. and Bilello, M. and Calabrese, E. and others},
  journal = {arXiv preprint arXiv:2107.02314},
  year    = {2021},
  doi     = {10.48550/arXiv.2107.02314}
}
\end{document}